\documentclass{article}
\usepackage{ijcai18}

\usepackage{times}
\usepackage{xcolor}
\usepackage{soul}
\usepackage[utf8]{inputenc}
\usepackage[small]{caption}

\usepackage{amsmath}
\usepackage{algorithm,algorithmic}
\usepackage{subfigure}
\usepackage{graphicx}
\usepackage{mathtools}

\title{Exploration by Distributional Reinforcement Learning}

\author{
Yunhao Tang, 
Shipra Agrawal
\\ 
Columbia University IEOR\\
yt2541@columbia.edu,
sa3305@columbia.edu
}

\begin{document}

\maketitle

\begin{abstract}
We propose a framework based on distributional reinforcement learning and recent attempts to  combine Bayesian parameter updates with deep reinforcement learning. We show that our proposed framework conceptually unifies multiple previous methods in exploration. We also derive a practical algorithm that achieves efficient exploration on challenging control tasks.
\end{abstract}

\section{Introduction}
Deep reinforcement learning (RL) has enjoyed numerous recent successes in various domains such as  video games and robotics control \cite{schulman2015,duanxi2016,levine2016}. Deep RL algorithms typically apply naive exploration strategies such as $\epsilon-$greedy \cite{mnih2013,timothy2016}. However, such myopic strategies cannot lead to systematic exploration in hard environments \cite{osband2017}. 

We provide an exploration algorithm based on distributional RL \cite{bellemare2017} and recent attempts to combine Bayesian parameter updates with deep reinforcement learning. We show that the proposed algorithm provides a conceptual unification of multiple previous methods on exploration in deep reinforcement learning setting. We also show that the algorithm achieves efficient exploration in challenging environments.

\section{Background}

\subsection{Markov Decision Process and Value Based Reinforcement Learning}
In a Markov Decision Process (MDP), at time step $t \geq 0$, an agent is in state $s_t \in \mathcal{S}$, takes action $a_t \in \mathcal{A}$, receives reward $r_t$ and gets transitioned to next state $s_{t+1} \sim p(s_{t+1}|s_t,a_t)$. At time $t = 0$ the agent's state distribution follows $s_0 \sim \rho(s_0)$. A policy is a mapping from a state to a distribution over action $a_t \sim \pi(\cdot|s_t)$. The objective is to find a policy $\pi$ to maximize the discounted cumulative reward 
\begin{align}
J = E_{s_0 \sim \rho, a_t \sim \pi(\cdot|s_t)} \big[ \sum_{t=0}^\infty r_t \gamma^t \big],
\label{eq:reward}
\end{align}
where $\gamma \in (0,1]$ is a discount factor. In state $s$, the action-value function $Q^\pi(s,a)$ is defined as the expected cumulative reward that could be received by first taking action $a$ and following policy $\pi$ thereafter
$$Q^\pi (s,a) = E_{a_t \sim \pi(\cdot|s_t)} \big[ \sum_{t=0}^\infty r_t \gamma^t | s_0 = s, a_0 = a \big].$$
From the above definition, it can be shown that $Q^\pi(s_t,a_t)$ satisfies the Bellman equation
$$Q^\pi(s_t,a_t) = E[r_t + \gamma Q^\pi(s_{t+1},a_{t+1})], \ \forall (s_t,a_t).$$
Let $\pi^\ast = \arg\max_\pi J$ be the optimal policy and $Q^\ast(s,a)$ its action value function. $Q^\ast(s,a)$ satisfies the following Bellman equation
$$Q^\ast(s_t,a_t) = E \big[ r_t + \gamma \max_a Q^\ast(s_{t+1},a)\big],\ \forall (s_t,a_t).$$
The above equations illustrate the temporal consistency of the action value functions that allows for the design of learning algorithms. Define Bellman operator
$$\mathcal{T}^\ast Q(s_t,a_t) \coloneqq E[r_t + \gamma \max_{a^\prime} Q(s_{t+1},a^\prime)].$$
When $\gamma \in (0,1)$, starting from any $Q^{(0)}(s,a)$, iteratively applying the operator $Q^{(t+1)}(s,a) \leftarrow T^\ast Q^{(t)}(s,a)$ leads to convergence $Q^{(t)}(s,a) \rightarrow Q^\ast(s,a)$ as $t\rightarrow \infty$.

In high dimensional cases, it is critical to use function approximation as a compact representation of action values. Let $Q_\theta(s,a)$ be such a function with parameter $\theta$ that approximates a table of action values with entry $(s,a)$. The aim is to find $\theta$ such that $Q_\theta(s,a) \approx Q^\ast(s,a)$. Let $\Pi$ be the operator that  projects arbitrary vector $Q(s,a) \in R^{|\mathcal{S}|\times |\mathcal{A}|}$ to the subspace spanned by function $Q_\theta(s,a)$. Since the update of action values can now only take place in the subspace spanned by function $Q_\theta(s,a)$, the iterate $Q^{(t)}(s,a)$ is updated as $Q^{(t+1)}(s,a) \leftarrow \Pi \mathcal{T}^\ast Q^{(t)}(s,a)$.
In cases where $Q_\theta(s,a)$ is linear, the above procedure can be shown to converge \cite{tsitsiklis1996}. However, in cases where $Q_\theta(s,a)$ is nonlinear (neural network), the function approximation becomes more expressive at the cost of no convergence guarantee. Many deep RL algorithms are designed following the above formulation, such as Deep Q Network (DQN) \cite{mnih2013}.

\subsection{Distributional Reinforcement Learning}
Following \cite{bellemare2017}, instead of considering action value $Q^\pi(s,a)$ under policy $\pi$, which is itself an expectation, consider the random return at $(s_t,a_t)$ by following policy $\pi$, $R^\pi(s_t,a_t) = \sum_{t^\prime \geq t} r_{t^\prime} \gamma^{t^\prime-t}$. It follows that $Q^\pi(s,a) = E[R^\pi(s,a)]$. Let $Z^\pi(s,a)$ be the distribution of $R^\pi(s,a)$. The Bellman equation for random return is similar to that of the action value functions $$Z^\pi(s_t,a_t) =_D r_t + \gamma Z^\pi(s_{t+1},\pi(s_{t+1})),$$
where both sides are distributions and $=_D$ denotes equality in distribution.\footnote{In future notations, we replace $=_D$ by $=$ for simplicity.} Define distributional Bellman operator $\mathcal{H}^\pi$ under policy $\pi$ as 
$$\mathcal{H}^\pi Z(s_t,a_t) \coloneqq r_t + \gamma Z(s_{t+1},\pi(s_{t+1})).$$
Notice that $\mathcal{H}^\pi$ operates on distributions. Define $\mathcal{H}^\ast$ as follows
$$\mathcal{H}^\ast Z \coloneqq \mathcal{H}^{\pi^\ast} Z,\ \  \text{for some optimal policy} \ \pi^\ast .$$
When $\gamma \in (0,1)$, starting from any distribution $Z^{(0)}(s,a)$, applying the operator as $Z^{(t+1)}(s,a) \leftarrow \mathcal{H}^\ast Z^{(t)}(s,a)$ leads to convergence in expectation $E[Z^{(t)}(s,a)] \rightarrow Q^\ast(s,a)$. However, the distribution $Z^{(t)}(s,a)$ itself may not weakly converge. 

To design a practical algorithm, one must use a parametric family of distribution $Z_\theta(s,a)$ to approximate $Z^\pi(s,a)$, with parameter $\theta$. Let $D(Z_1,Z_2)$ be a discrepancy measure between distribution $Z_1$ and $Z_2$. Define the projection operator $\Pi$ as follows
$$\Pi Z \coloneqq \arg\min_{Z_\theta} D(Z,Z_\theta).$$
In other words, $\Pi$ projects a distribution $Z$ into another distribution $Z_\theta$ in the parametric family with smallest discrepancy from $Z$. Hence the distribution $Z$ is updated as
$Z^{(t+1)}(s,a) \leftarrow \Pi \mathcal{H}^\ast Z^{(t)}(s,a)$.
In practice, the operator is applied to different entries $(s,a)$ asynchronously. For a given pair $(s_t,a_t)$, one first selects a greedy action for next state
$$a^\prime = \arg\max_a E[ Z_\theta(s_{t+1},a)],$$
then updates the distribution $Z_\theta(s_t,a_t)$ to match the target distribution by minimizing the discrepancy
\begin{align}
\min_\theta D(r_t + \gamma Z_\theta(s_{t+1},a^\prime),Z_\theta(s_t,a_t)).
\label{eq:minkl}
\end{align}
When only samples $x_i \sim r_t + \gamma Z_\theta(s_{t+1},a^\prime), 1\leq i\leq N$ are available, let the empirical distribution be $\hat{Z} = \sum_{i=1}^N \frac{1}{N}\delta(x-x_i)$ \footnote{$\delta(x-x_i)$ is the Dirac distribution that assigns point mass of probability $1$ at $x=x_i$.}, then (\ref{eq:minkl}) reduces to minimizing
$D(\hat{Z},Z_\theta)$. 

\section{Related Work}
In reinforcement learning (RL), naive explorations such as $\epsilon-$greedy \cite{mnih2013,timothy2016} do not explore well because local perturbations of actions break the consistency between consecutive steps \cite{osband2015}. A number of prior works apply randomization to parameter space \cite{fortunato2017,plappert2016} to preserve the consistency in exploration, but their formulations are built on heuristics. Posterior sampling is a principled exploration strategy in the bandit setting \cite{william1933,russo2017}, yet its extension to RL  \cite{osband2013} is hard to scale to large problems. More recent prior works have formulated the exploration strategy as sampling randomized value functions and interpreted the algorithm as approximate posterior sampling \cite{osband2016,osband2017}. Instead of modeling value functions, our formulation is built on modeling return distributions which reduces to exact posterior sampling in the bandit setting.

Following similar ideas of randomized value function, multiple recent works have combined approximate Bayesian inference \cite{blei2015,blei2017} with Q learning and justified the efficiency of  exploration by relating to posterior sampling \cite{lipton2016,tang2017,kamyar2017,thomas2017}. Though their formulations are based on randomized value functions, we offer an alternate interpretation by modeling return distribution and provide a conceptual framework that unifies these previous methods (Section 5). We will also provide a potential approach that extends the current framework to policy based methods as in \cite{henderson2017}.

Modeling return distribution dated back to early work of \cite{dearden1998,morimura2010,morimura2012},  where learning a return distribution instead of only its expectation presents a more statistically challenging task but provides more information during control. More recently, \cite{bellemare2017} applies a histogram to learn the return distribution and displays big performance gains over DQN \cite{mnih2013}. Based on \cite{bellemare2017}, we provide a more general distributional learning paradigm that combines return distribution learning and exploration based on approximate posterior sampling.

\section{Exploration by Distributional Reinforcement Learning}
\subsection{Formulation}
Recall that $Z(s,a)$ is the return distribution for state action pair $(s,a)$. In practice, we approximate such distribution by a parametric distribution $Z_\theta(s,a)$ with parameter $\theta$. Following \cite{bellemare2017}, we take the discrepancy to be KL divergence. Recall $\hat{Z}$ is the empirical distribution of samples $\hat{Z} = \sum_{i=1}^N \frac{1}{N}\delta(x-x_i)$, hence the KL divergence reduces to
\begin{align}
KL[\hat{Z} || Z_\theta] = \sum_{i=1}^N \frac{1}{N} \log \frac{\frac{1}{N}}{Z_\theta(x_i)} = -\frac{1}{N}\sum_{i=1}^N \log Z_\theta(x_i),
\label{eq:discrepancy}
\end{align}
where we have dropped a constant $- \log N$ in the last equality. Let $\theta$ follow a given distribution $\theta\sim q_\phi(\theta)$ with parameter $\phi$. We propose to minimize the following objective
\begin{align}
\min_\phi  E_{\theta\sim q_\phi(\theta)}[-\sum_{i=1}^N \log Z_\theta(x_i)] - H(q_\phi(\theta)),
\label{eq:variationalobj}
\end{align}
where $H(q_\phi(\theta))$ is the entropy of $q_\phi(\theta)$. Note that (\ref{eq:discrepancy}) corresponds to the projection step $\Pi\mathcal{H}^\ast$ defined in (\ref{eq:minkl}), and the first term of (\ref{eq:variationalobj}) takes an expectation of projection discrepancy over the distribution $\theta \sim q_\phi(\theta)$. The intuition behind (\ref{eq:variationalobj}) is that by the first term, the objective encourages low expected discrepancy (which is equivalent to Bellman error) to learn optimal policies; the second term serves as an exploration bonus to encourage a dispersed distribution over $\theta$ for better exploration during learning.

We now draw the connection between (\ref{eq:variationalobj}) and approximate Bayesian inference. First assign an improper uniform prior on $\theta$, i.e. $p(\theta) \propto 1$. The posterior is defined by Bayes rule given the data $\{x_i\}_{i=1}^N$ as $p(\theta|\{x_i\}_{i=1}^N) \propto p(\theta)p(\{x_i\}_{i=1}^N | \theta)$
where $p(\{x_i\}_{i=1}^N|\theta) = \Pi_i p(x_i|\theta)$ \footnote{We assume samples drawn from the next state distributions are i.i.d. as in \cite{bellemare2017}.}. Since by definition $p(x_i|\theta) = Z_\theta(x_i)$, (\ref{eq:variationalobj}) is equivalent to 
\begin{align}
\min_\phi KL[q_\phi(\theta)||p(\theta|\{x_i\}_{i=1}^N)].
\label{eq:kl}
\end{align}
Hence to minimize the objective ($\ref{eq:variationalobj}$) is to search for a parametric distribution $q_\phi(\theta)$ to approximate the posterior $p(\theta|\{x_i\}_{i=1}^N)$. From (\ref{eq:kl}) we can see that the posterior $p(\theta|\{x_i\}_{i=1}^N)$ is the minimizer policy of (\ref{eq:variationalobj}), which achieves the optimal balance between minimizing low discrepancy and being as random as possible. The close resemblance between our formulation and posterior sampling partially justifies the potential strength of our exploration strategy. 

\subsection{Generic Algorithm}
A generic algorithm \textbf{Algorithm 1} can be derived from (\ref{eq:kl}). We start with a proposed distribution $q_\phi(\theta)$ over parameter $\theta$ and a distribution model $Z_\theta(s,a)$. During control, in state $s_t$, we sample a parameter from $\theta \sim q_\phi(\theta)$ and choose action $a_t = \arg\max_a E[Z_\theta(s_t,a)]$. This is equivalent to taking an action based on the approximate posterior probability that it is optimal. During training, we sample from one-step lookahead distribution of the greedy action, and update parameter by optimizing (\ref{eq:variationalobj}). 
\begin{algorithm}[H]
	\begin{algorithmic}[1]
		\STATE INPUT:  generic return distribution $Z_\theta(s,a)$ with parameter $\theta$, parameter distribution $q_\phi(\theta)$ with parameter $\phi$.

		\WHILE {not converged}	
		\STATE \textbf{\emph{// Control}}
		\STATE Sample $\theta \sim q_\phi(\theta)$.
		\STATE In state $s_t$, choose $a_t = \arg\max_a E[Z_\theta(s_t,a)]$, get transition $s_{t+1}$ and reward $r_t$.
		\STATE \textbf{\emph{// Training}}
		\STATE Given state action pair $s_t,a_t$, choose greedy one-step lookahead distribution
		$a^\prime = \arg\max_a E[r_t + \gamma Z_\theta(s_{t+1},a)]$.
		\STATE Sample from the distribution $r_t + \gamma Z_\theta(s_{t+1},a^\prime)$ and let $\hat{Z}$ be the empirical distribution of samples, update parameter $\phi$ by minimizing objective (\ref{eq:variationalobj}).
		\ENDWHILE
	\end{algorithmic}
	\caption{Exploration by Distributional RL: Generic}
\end{algorithm}

\subsection{Practical Algorithm: Gaussian Assumption}
We turn \textbf{Algorithm 1} into a practical algorithm by imposing assumption on $Z_\theta(s,a)$. \cite{dearden1998} assumes $Z_\theta(s,a)$ to be Gaussian based on the assumption that the chain is ergodic and $\gamma$ close to $1$. We make this assumption here and let $Z_\theta(s,a)$ be a Gaussian with parametrized mean $Q_\theta(s,a)$ and fixed standard error $\sigma$.  The objective (\ref{eq:discrepancy}) reduces to
\begin{align}
\min_\theta \frac{1}{N} \sum_{i=1}^N \frac{(Q_\theta(s,a) - x_i)^2}{2\sigma^2}.
\label{eq:bellman}
\end{align}
We now have an analytical form $E[Z_\theta(s,a)] = Q_\theta(s,a)$. The objective (\ref{eq:variationalobj}) reduces to
\begin{align}
\min_\phi  E_{\theta\sim q_\phi(\theta)}[\sum_{i=1}^N \frac{(Q_\theta(s,a) - x_i)^2}{2\sigma^2}] - H(q_\phi(\theta)).
\label{eq:Gauss}
\end{align}

\begin{algorithm}[H]
	\begin{algorithmic}[1]
		\STATE INPUT:  target parameter update period $\tau$ ; learning rate $\alpha$; Gaussian distribution parameter $\sigma^2$.
		\STATE INITIALIZE: parameters $\phi,\phi^{-}$; replay buffer $B \leftarrow \{\}$; step counter $counter \leftarrow 0$.
		\FOR {$e=1,2,3...E$}
		\WHILE {episode not terminated}
		\STATE $counter \leftarrow counter + 1$.
		\STATE Sample $\theta \sim q_\phi(\theta)$.
		\STATE In state $s_t$, choose $a_t = \arg\max_a Q_\theta(s_t,a)$, get transition $s_{t+1}$ and reward $r_t$.
		\STATE Save experience tuple $\{s_t,a_t,r_t,s_{t+1}\}$ to buffer $B$.
		\STATE Sample $N$ parameters $\theta_j^{-} \sim q_{\phi^-}(\theta^{-})$ and sample $N$ tuples $D = \{s_j,a_j,r_j,s_j^\prime\}$ from $B$.
		\STATE Sample target $x_j \sim  r_j + \gamma Z_{\theta_j^{-}}(s_j^\prime,a^\prime)$ for $j$th tuple in $D$ where $a^\prime$ is greedy w.r.t. $Q_{\theta^{-}}(s_j^\prime,a)$.
		\STATE Take gradient $\Delta \phi$ of the KL divergence in (\ref{eq:Gauss}). 
		\STATE $\phi\leftarrow \phi - \alpha \Delta \phi$.
		\IF {$counter \ \text{mod}\  \tau = 0$}
		\STATE Update target parameter $\phi^{-} \leftarrow \phi$.
		\ENDIF
		\ENDWHILE
		\ENDFOR
	\end{algorithmic}
	\caption{Exploration by Distributional RL: Gaussian}
\end{algorithm}

Parallel to the principal network $q_\phi(\theta)$ with parameter $\phi$, we maintain a target network $q_{\phi^-}(\theta^-)$ with parameter $\phi^-$ to stabilize learning \cite{mnih2013}. Samples for updates are generated by target network $\theta^- \sim q_{\phi^-}(\theta^-)$. We also maintain a replay buffer $B$ to store off-policy data.

\subsection{Randomized Value Function as Randomized Critic for Policy Gradient}
In off-policy optimization algorithm like Deep Deterministic Policy Gradient (DDPG) \cite{timothy2016}, a policy $\pi_{\theta_p}(s)$ with parameter $\theta_p$ and a critic $Q_\theta(s,a)$ with parameter $\theta$ are trained at the same time. The policy gradient of reward objective (\ref{eq:reward}) is 
\begin{align}
\nabla_{\theta_p} J &= E[\sum_{t} \nabla_a Q^\pi(s,a) |_{a = \pi_{\theta_p}(s_t)} \nabla_{\theta_p} \pi_{\theta_p}(s_t)] \nonumber \\
&\approx E[\sum_{t} \nabla_a Q_\theta(s,a) |_{a = \pi_{\theta_p}(s_t)} \nabla_{\theta_p} \pi_{\theta_p}(s_t)],
\label{eq:ddpggrad}
\end{align}
where replacing true $Q^\pi(s,a)$ by a critic $Q_\theta(s,a)$ introduces bias but largely reduces variance \cite{timothy2016}.

To extend the formulation of \textbf{Algorithm 2} to policy based methods, we can interpret $Q_\theta(s,a)$ as a randomized critic with a distribution induced by $\theta \sim q_\phi(\theta)$. At each update we sample a parameter $\hat{\theta} \sim q_\phi(\theta)$ and compute the policy gradient (\ref{eq:ddpggrad}) through the sampled critic $Q_{\hat{\theta}}(s,a)$ to update $\theta_p$. The distributional parameters $\phi$ are updated as in \textbf{Algorithm 2} with the greedy actions replaced by actions produced by the policy $\pi_{\theta_p}$.

Policy gradients computed from randomized critic may lead to better exploration directly in the policy space as in \cite{plappert2016}, since the uncertainties in the value function can be propagated into the policy via gradients through the uncertain value functions.

\section{Connections with Previous Methods}
We now argue that the above formulation provides a conceptual unification to multiple previous methods. We can recover the same objective functions as previous methods by properly choosing the parametric form of return distribution $Z_\theta(s,a)$, the distribution over model parameter $q_\phi(\theta)$ and the algorithm to optimize the objective (\ref{eq:kl}).
\subsection{Posterior Sampling for Bandits}
In the bandit setting, we only have a set of actions $a\in \mathcal{A}$. Assume the underlying reward for each action $a$ is Gaussian distributed. To model the return distribution of action $a$, we set $Z_\theta(a)$ to be Gaussian with unknown mean parameters $\mu_a$, i.e. $Z_\theta = \mathcal{N}(\mu_a,\sigma^2)$. We assume the distribution over parameter $q_\phi(\mu)$ to be Gaussian as well. Due to the conjugacy between improper uniform prior $p(\mu)$ (assumed in Section 4.1) and likelihood $Z_\theta(a)$, the posterior $p(\mu|\{x_i\})$ is still Gaussian. We can minimize (\ref{eq:kl}) exactly by setting $q_\phi(\mu) = p(\mu|\{x_i\})$. During control, \textbf{Algorithm 1} selects action $a_t = \arg\max_a \mu(a)$ with sampled $\mu\sim q_\phi(\mu)=p(\mu|\{x_i\})$, which is exact posterior sampling. This shows that our proposed algorithm reduces to exact posterior sampling for bandits. For general RL cases, the equivalence is not exact but this connection partially justifies that our algorithm can achieve very efficient exploration.

\subsection{Deep Q Network with Bayesian Updates}
Despite minor algorithmic differences, \textbf{Algorithm 2} has very similar objective as Variational DQN \cite{tang2017}, BBQ Network \cite{lipton2016} and Bayesian DQN \cite{kamyar2017}, i.e. all three algorithms can be interpreted as having Gaussian assumption over return distribution $Z_\theta(s,a)\sim \mathcal{N}(Q_\theta(s,a),\sigma^2)$ and proposing Gaussian distribution over parameters $q_\phi(\theta)$. 
However, it is worth recalling that \textbf{Algorithm 2} is formulated by modeling return distributions, while previous methods are formulated by randomizing value functions. 

If we are to interpret these three algorithms as instantiations of \textbf{Algorithm 2}, the difference lies in how they optimize (\ref{eq:Gauss}). Variational DQN and BBQ apply variational inference to minimize the divergence between $q_\phi(\theta)$ and posterior $p(\theta|\{x_i\})$, while Bayesian DQN applies exact analytical updates (exact minimization of (\ref{eq:Gauss})), by using the conjugacy of prior and likelihood distributions as discussed above. \textbf{Algorithm 1} generalizes these variants of DQN with Bayesian updates by allowing for other parametric likelihood models $Z_\theta(s,a)$, though in practice Gaussian distribution is very popular due to its simple analytical form. 

To recover NoisyNet \cite{fortunato2017} from (\ref{eq:Gauss}), we can properly scale the objective (by multiplying (\ref{eq:Gauss}) by $\sigma^2$) and let $\sigma \rightarrow 0$. This implies that NoisyNet makes less strict assumption on return distribution (Gauss parameter $\sigma$ does not appear in objective) but does not explicitly encourage exploration by adding entropy bonus, hence the exploration purely relies on the randomization of parameter $\theta$. To further recover the objective of DQN \cite{mnih2013}, we set $q_\phi(\theta) = \delta(\theta - \phi)$ to be the Dirac distribution. Finally, since DQN has no randomness in the parameter $\theta$, its exploration relies on greedy action perturbations.

\subsection{Distributional RL}
Distributional RL \cite{bellemare2017} models return distribution using categorical distribution and does not introduce parameter uncertainties. Since there is no distribution over parameter $\theta$, \textbf{Algorithm 1} recovers the exact objective of distributional RL from (\ref{eq:variationalobj}) by setting $q_\phi(\theta) = \delta(\theta-\phi)$ and letting $Z_\theta(s,a)$ be categorical distributions. As the number of atoms in the categorical distribution increases, the modeling becomes increasingly close to non-parametric estimation. Though having more atoms makes the parametric distribution more expressive, it also poses a bigger statistical challenge during learning due to a larger number of parameters. As with general $Z_\theta(s,a)$, choosing a parametric form with appropriate representation power is critical for learning.

\section{Experiments}
In all experiments, we implement \textbf{Algorithm 2} and refer to it as GE (Gauss exploration) in the following. We aim to answer the following questions,
\begin{itemize}
\item In environments that require consistent exploration, does GE achieve more efficient exploration than conventional naive exploration strategies like $\epsilon-$greedy in DQN and direct parameter randomization in NoisyNet?
\item When a deterministic critic in an off-policy algorithm like DDPG \cite{timothy2016} is replaced by a randomized critic, does the algorithm achieve better exploration?
\end{itemize}

\subsection{Testing Environment}
\paragraph{Chain MDP.} The chain MDP \cite{osband2016} (Figure 1) serves as a benchmark to test if an algorithm entails consistent exploration. The environment consists of $N$ states and each episode lasts $N+9$ time steps. The agent has two actions $\{\text{left},\text{right}\}$ at each state $s_i,1\leq i\leq N$, while state $s_1,s_N$ are both absorbing. The transition is deterministic. At state $s_1$ the agent receives reward $r = \frac{1}{1000}$, at state $s_N$ the agent receives reward $r = 1$ and no reward anywhere else. The initial state is always $s_2$, making it hard for the agent to escape local optimality at $s_1$.  If the agent explores uniformly randomly, the expected number of time steps required to reach $s_N$ is $2^{N-2}$. For large $N$, it is almost not possible for the randomly exploring agent to reach $s_N$ in a single episode, and the optimal strategy to reach $s_N$ will never be learned. 
\begin{figure}[t]
\centering
\includegraphics[width=1.05\linewidth]{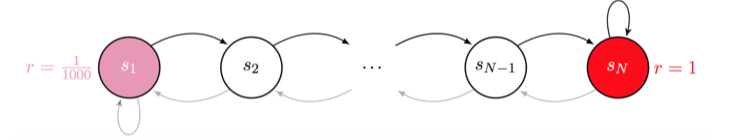}
\caption{Chain MDP with $N$ states}
\end{figure}
\paragraph{Sparse Reward Environments.} All RL agents require reward signals to learn good policies. In sparse reward environment,  agents with naive exploration strategies randomly stumble around for most of the time and require many more samples to learn good policies than agents that explore consistently. We modify the reward signals in OpenAI gym \cite{brockman2016} and MuJoCo benchmark tasks \cite{todorov2012} to be sparse as follows.
\begin{itemize}
\item MountainCar, Acrobot: $r = 1$ when the episode terminates and $r = 0$ otherwise.
\item CartPole, InvertedPendulum, InvertedDoublePendulum: $r = -1$ when the episode terminates and $r = 0$ otherwise.
\end{itemize}

\subsection{Experiment Results}

\paragraph{Exploration in Chain MDP.} In Figure 2 (a) - (c) we compare DQN vs NoisyNet vs GE in Chain MDP environments with different number of states $N$. When $N = 10$, all three algorithms can solve the task. When $N = 50$, DQN cannot explore properly and cannot make progress, GE explores more efficiently and converges to optimal policy faster than NoisyNet. When $N = 100$, both NoisyNet and DQN get stuck while GE makes progress more consistently. Compared to Bootstrapped DQN (BDQN)\cite{osband2016}, GE has a higher variance when $N=100$. This might be because BDQN represents the distribution using multiple heads and can approximate more complex distributions, enabling better exploration on this particular task. In general, however, our algorithm is much more computationally feasible than BDQN yet still achieves very efficient exploration.

Figure 2 (d) plots the state visit frequency for GE vs. DQN within the \textbf{first 10 episodes} of training. DQN mostly visits states near $s_2$ (the initial state), while GE visits a much wider range of states. Such active exploration allows the agent to consistently visit $s_N$ and learns the optimal policy within a small number of iterations.
\begin{figure}[t]
\centering
\subfigure[Chain MDP $N = 10$]{\includegraphics[width=.45\linewidth]{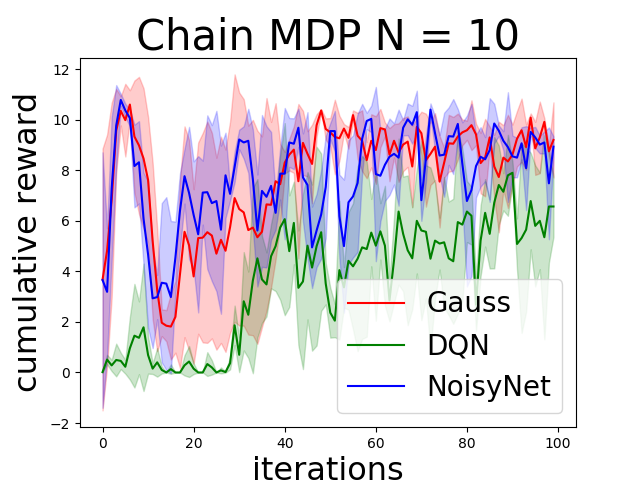}}
\subfigure[Chain MDP $N = 50$]{\includegraphics[width=.45\linewidth]{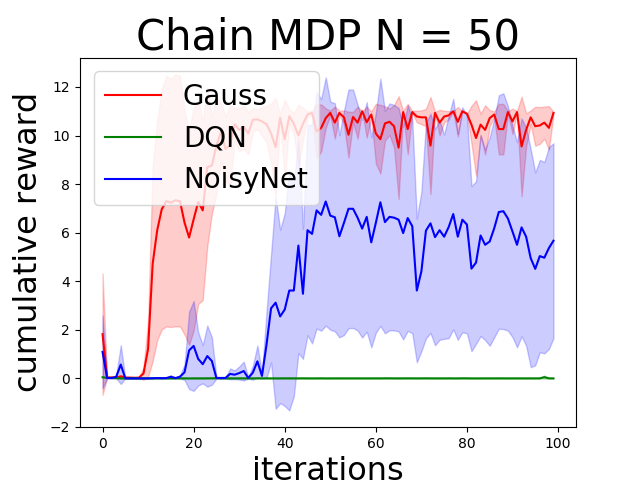}}
\subfigure[Chain MDP $N = 100$]{\includegraphics[width=.45\linewidth]{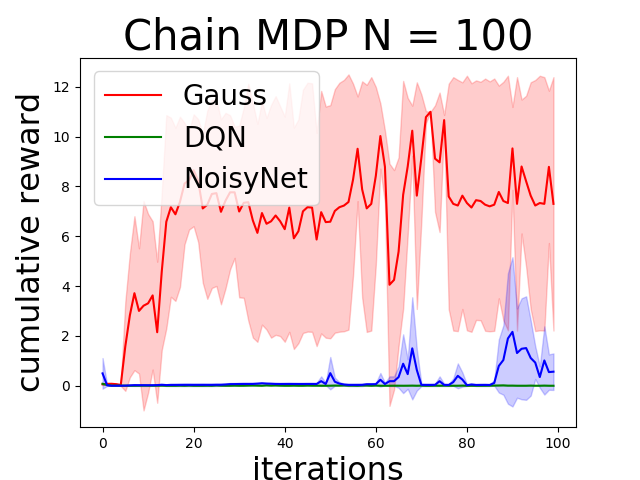}}
\subfigure[State visit frequency]{\includegraphics[width=.45\linewidth]{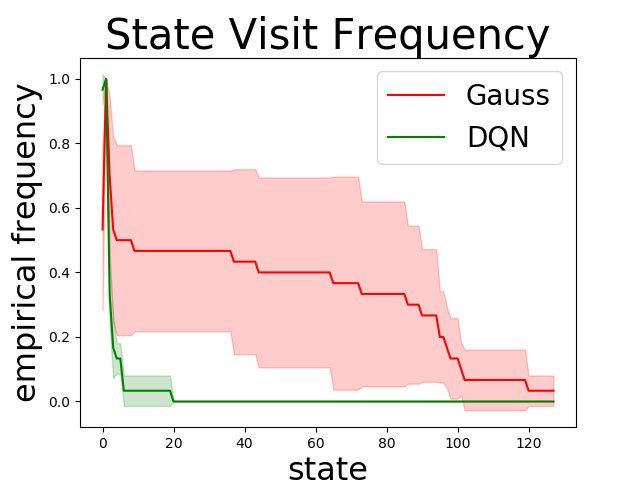}}
\caption{\small{Comparison of DQN vs NoisyNet vs GE on Chain MDP environments with (a) $N = 10$ (b) $N = 50$ and (c) $N = 100$ states. Figure 2 (d) plots state visit frequency within the \textbf{first iteration} in training for Gauss vs. DQN in Chain MDP $N = 128$. For state $s_i$, set $c_i = 1$ if $s_i$ is ever visited in one episode and $c_i = 0$ otherwise. The moving average of $c_i$ across multiple episodes computes the state visit frequency. Each iteration consists of $20$ episodes.}}
\end{figure}
\paragraph{Exploration in Sparse Reward Environments.}
In Figure 3  (a) - (c) we present the comparison of three algorithms in sparse reward environments. For each environment, we plot the rewards at a different scale. In CartPole, the plotted cumulative reward is the episode length; in MountainCar, the plotted cumulative reward is $1$ for reaching the target within one episode and $0$ otherwise; in Acrobot, the plotted cumulative reward is the negative of the episode length. In all sparse reward tasks, GE entails much faster progress than the other two algorithms. For example, in Sparse MountainCar, within the given number of iterations, DQN and NoisyNet have never (or very rarely) reached the target, hence they make no (little) progress in cumulative reward. On the other hand, GE reaches the targets more frequently since early stage of the training, and makes progress more steadily.

In Figure 3 (d) we plot the state visit trajectories of GE vs. DQN in Sparse MountainCar. The vertical and horizontal axes of the plot correspond to two coordinates of the state space. Two panels of (d) correspond to training after $10$ and $30$ iterations respectively. As the training proceeds, the state visits of DQN increasingly cluster on a small region in state space and fail to efficiently explore. On the contrary, GE maintains a widespread distribution over states and can explore more systematically.
\begin{figure}[t]
\centering
\subfigure[Sparse CartPole]{\includegraphics[width=.45\linewidth]{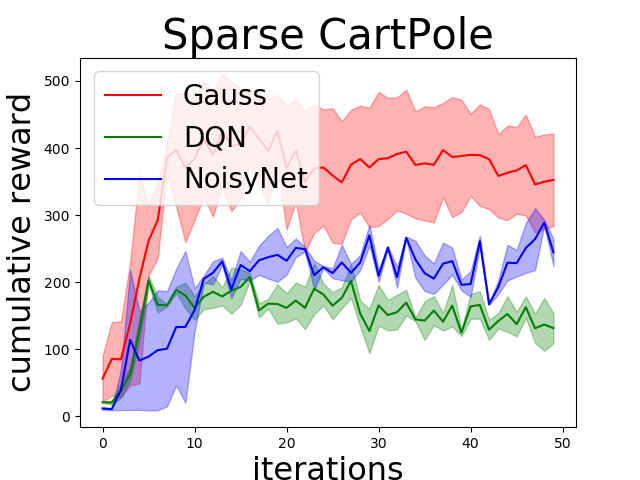}}
\subfigure[Sparse MountainCar]{\includegraphics[width=.45\linewidth]{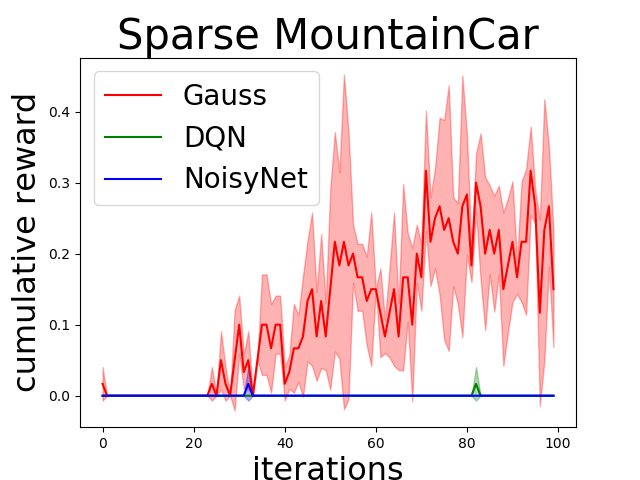}}
\subfigure[Sparse Acrobot]{\includegraphics[width=.45\linewidth]{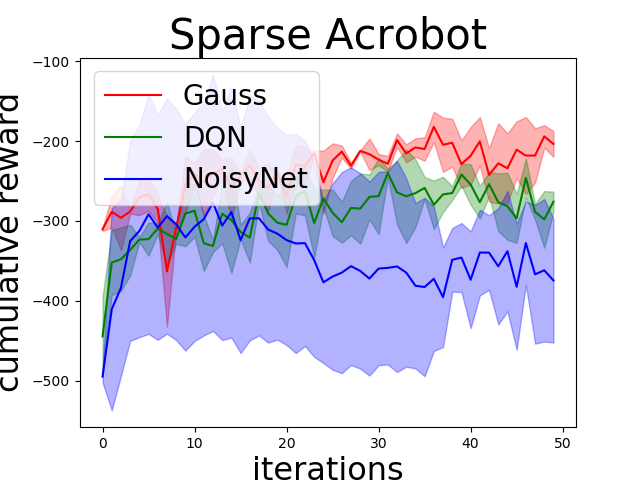}}
\subfigure[State visit trajectories]{\includegraphics[width=.45\linewidth]{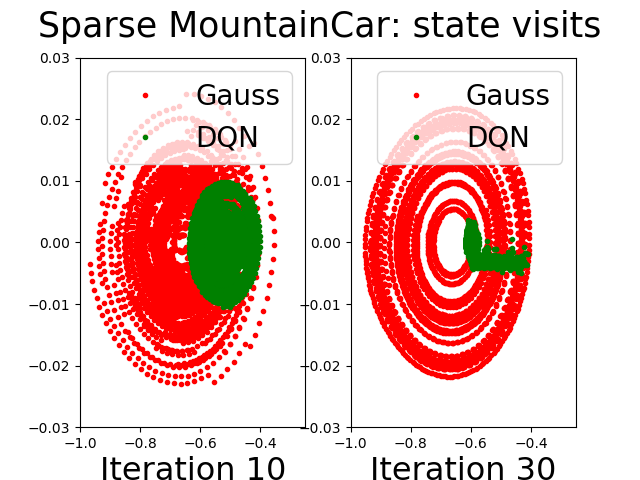}}
\caption{Comparison of DQN vs NoisyNet vs GE on sparse reward environments (a) Sparse CartPole (b) Sparse MountainCar (c) Sparse Acrobot. Each iteration corresponds to $20$ episodes. Rewards are plotted using moving windows of $20$ episodes. Figure 3 (d) plots state visit trajectories for Gauss vs. DQN in Sparse MountainCar. Left panel of (d) is training after $10$ iterations and the right panel is after $30$ iterations. The vertical and horizontal axes correspond to two coordinates of the state space. }
\end{figure}
\paragraph{Randomized Critic for Exploration.}
We evaluate the performance of DDPG with different critics. When DQN is used as a critic, the agent explores by injecting noise into actions produced by the policy \cite{timothy2016}. When critics are NoisyNet or randomized DQN with GE, the agent explores by updating its parameters using policy gradients computed through randomized critics, effectively injecting noise into the parameter space. In conventional continuous control tasks (Figure 4 (a) and (b)), randomized critics do not enjoy much advantage: for example, in simple control task like InvertedPendulum, where exploration is not important, DDPG with action noise injection makes progress much faster (Figure 4 (a)), though DDPG with randomized critics seem to make progress in a steadier manner. In sparse reward environments (Figure 4 (c) and (d)), however, DDPG with randomized critics tend to make progress at a slightly higher rate than action noise injection.

\begin{figure}[t]
\centering
\subfigure[InvertedPendulum]{\includegraphics[width=.45\linewidth]{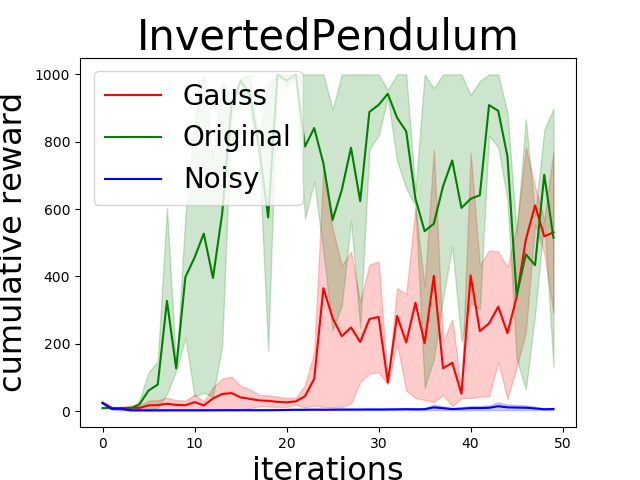}}
\subfigure[InvertedDoublePendulum]{\includegraphics[width=.45\linewidth]{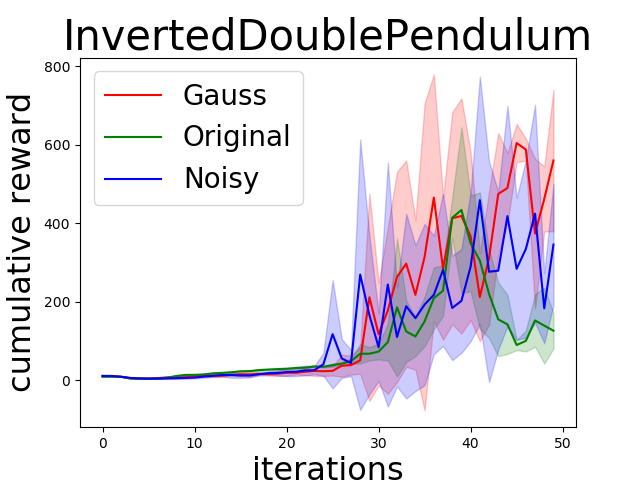}}
\subfigure[Sparse InvertedPendulum]{\includegraphics[width=.45\linewidth]{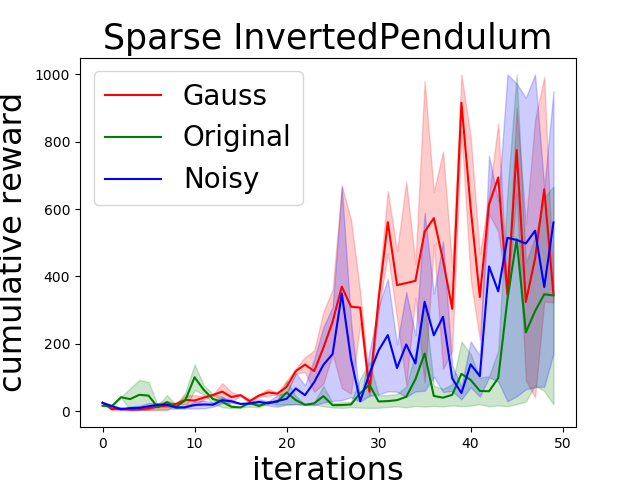}}
\subfigure[Sparse DoublePendulum]{\includegraphics[width=.45\linewidth]{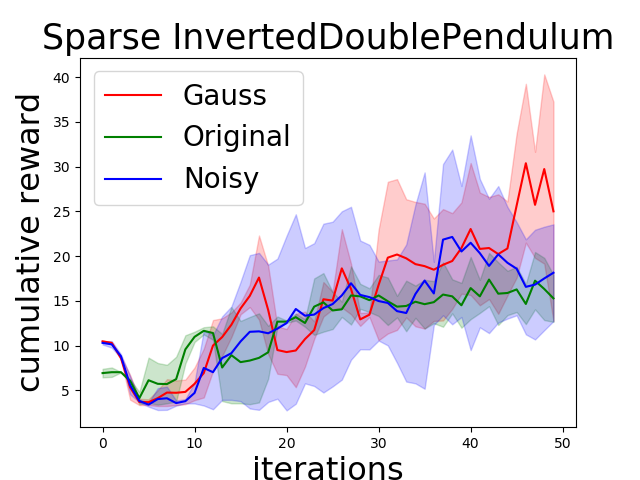}}
\caption{Comparison of original Q function (DQN) vs NoisyNet vs GE \textbf{as baselines for DDPG} on sparse reward environments (a) InvertedPendulum (b) InvertedDoublePendulum (c) Sparse InvertedPendulum (d) Sparse InvertedDoublePendulum.}
\end{figure}

\paragraph{Hyper-parameter.}
In all experiments, we set $q_\phi(\theta)$ to be factorized Gaussian. In GE, as in NoisyNet \cite{fortunato2017}, each parameter $\theta$ in a fully connected layer (weight and bias) has two distributional parameters: the mean $\mu_\theta$ and standard error $\sigma_\theta$. Set $\sigma_\theta = \log (1 + \exp(-\rho_\theta))$ and let $\rho_\theta$ be the actual hyper-parameter to tune. If $\rho_\theta$ is large, the distribution over $\theta$ is widespread and the agent can execute a larger range of policies before committing to a solution. For both NoisyNet and GE, we require all $\rho_\theta$ to be the same, denoted as $\rho$, and set the range $\rho \in [-1,-10]$ for grid search. A second hyper-parameter for GE is the Gauss parameter $\sigma^2$ to determine the balance between expected Bellman error and entropy in (\ref{eq:Gauss}). In our experiments, we tune $\sigma$ on the log scale $\log_{10}\sigma \in [-1,-8]$.

We empirically find that both $\rho,\sigma$ are critical to the performance of GE. For each algorithm, We use a fairly exhaustive grid search to obtain the best hyper-parameters. Each experiment is performed multiple times and the reward plots in Figure 2,3,4 are averaged over five different seeds. In Figure 5, we plot the performance of GE under different $\rho$ and $\sigma$ on Sparse CartPole. From Figure 5  we see that the performance is not monotonic in $\rho,\sigma$: large $\sigma$ (small $\rho$) generally leads to more active exploration but may hinder fast convergence, and vice versa. One must strike a proper balance between exploration and exploitation to obtain good performance. In DQN, we set the exploration constant to be $\epsilon = 0.1$. In all experiments, we tune the learning rate $\alpha \in \{10^{-3},10^{-4},10^{-5}\}$. 

\begin{figure}[t]
\centering
\subfigure[Hyper-parameter $\rho$]{\includegraphics[width=.45\linewidth]{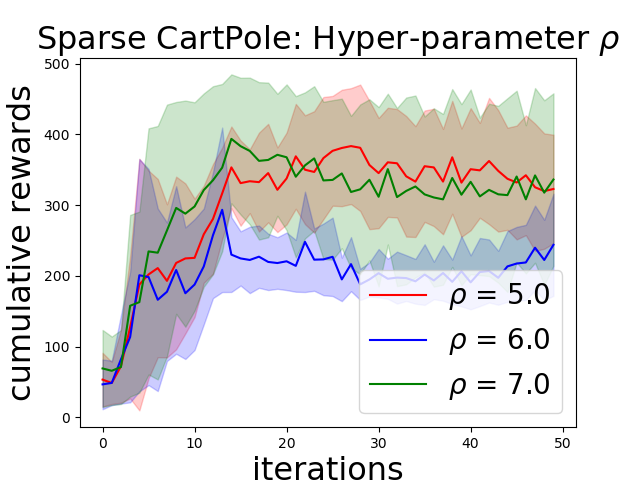}}
\subfigure[Hyper-parameter $\sigma$]{\includegraphics[width=.45\linewidth]{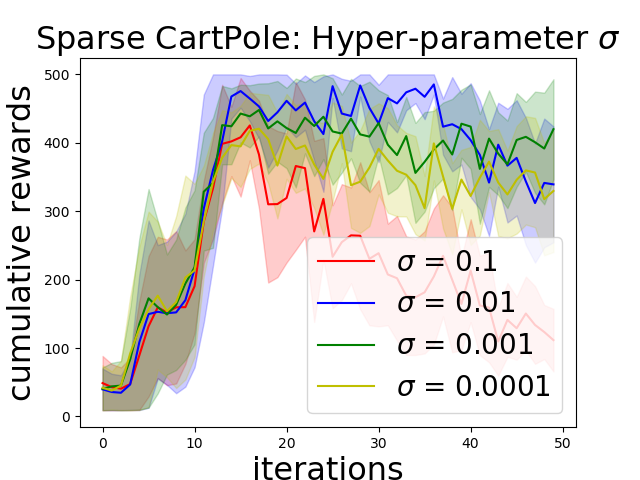}}
\caption{Hyper-parameter for Gauss Exploration (GE)}
\end{figure}

\section{Conclusion}
We have provided a framework based on distributional RL that unifies multiple previous methods on exploration in reinforcement learning, including posterior sampling for bandits as well as recent efforts in Bayesian updates of DQN parameters. We have also derived a practical algorithm based on the Gaussian assumption of return distribution, which allows for efficient control and parameter updates. We have observed that the proposed algorithm obtains good performance on challenging tasks that require consistent exploration. A further extension of our current algorithm is to relax the Gaussian assumption on return distributions. We leave it be future work if more flexible assumption can lead to better performance and whether it can be combined with model-based RL.

\appendix
\newpage
\bibliographystyle{named}
\bibliography{ijcai18}

\begin{thebibliography}{}

\bibitem[\protect\citeauthoryear{Azizzadenesheli \bgroup \em et al.\egroup
  }{2017}]{kamyar2017}
Kamyar Azizzadenesheli, Emma Brunskill, and Animashree Anandkumar.
\newblock Efficient exploration through bayesian deep q networks.
\newblock {\em Symposium on Deep Reinforcement Learning, NIPS}, 2017.

\bibitem[\protect\citeauthoryear{Bellemare \bgroup \em et al.\egroup
  }{2017}]{bellemare2017}
Marc~G. Bellemare, Will Dabney, and Remi Munos.
\newblock A distributional perspective on reinforcement learning.
\newblock {\em International Conference on Machine Learning}, 2017.

\bibitem[\protect\citeauthoryear{Blei \bgroup \em et al.\egroup
  }{2017}]{blei2017}
David~M. Blei, Alp Kucukelbir, and Jon~D. McAuliffe.
\newblock Variational inference: A review for statisticians.
\newblock {\em Journal of the American Statistical Association, Volume 112 -
  Issue 518}, 2017.

\bibitem[\protect\citeauthoryear{Brockman \bgroup \em et al.\egroup
  }{2016}]{brockman2016}
Greg Brockman, Vicki Cheung, Ludwig Pettersson, Jonas Schneider, John Schulman,
  Jie Tang, and Wojciech Zaremba.
\newblock Openai gym.
\newblock {\em Arxiv: 1606.01540}, 2016.

\bibitem[\protect\citeauthoryear{Dearden \bgroup \em et al.\egroup
  }{1998}]{dearden1998}
Richard Dearden, Nir Friedman, and Stuart Russel.
\newblock Bayesian q learning.
\newblock {\em American Association for Artificial Intelligence (AAAI)}, 1998.

\bibitem[\protect\citeauthoryear{Duan \bgroup \em et al.\egroup
  }{2016}]{duanxi2016}
Yan Duan, Xi~Chen, Rein Houthooft, John Schulman, and Pieter Abbeel.
\newblock Benchmarking deep reinforcement learning for continuous control.
\newblock {\em International Conference on Machine Learning}, 2016.

\bibitem[\protect\citeauthoryear{Fortunato \bgroup \em et al.\egroup
  }{2017}]{fortunato2017}
Meire Fortunato, Mohammad~Gheshlaghi Azar, Bilal Piot, Jacob Menick, Ian
  Osband, Alex Graves, Vlad Mnih, Remi Munos, Demis Hassabis, Ilivier Pietquin,
  Charles Blundell, and Shane Legg.
\newblock Noisy network for exploration.
\newblock {\em arXiv:1706.10295}, 2017.

\bibitem[\protect\citeauthoryear{Henderson \bgroup \em et al.\egroup
  }{2017}]{henderson2017}
Peter Henderson, Thang Doan, Riashat Islam, and David Meger.
\newblock Bayesian policy gradients via alpha divergence dropout inference.
\newblock {\em 2nd Workshop on Bayesian Deep Learning, NIPS}, 2017.

\bibitem[\protect\citeauthoryear{Levine \bgroup \em et al.\egroup
  }{2016}]{levine2016}
Sergey Levine, Chelsea Finn, Trevor Darrell, and Pieter Abbeel.
\newblock End to end training of deep visuomotor policies.
\newblock {\em Journal of Machine Learning Research}, 2016.

\bibitem[\protect\citeauthoryear{Lillicrap \bgroup \em et al.\egroup
  }{2016}]{timothy2016}
Timothy~P. Lillicrap, Jonathan~J. Hunt, Alexander Pritzel, Nicolas Heess, Tom
  Erez, Yuval Tassa, David Silver, and Daan Wierstra.
\newblock Continuous control with deep reinforcement learning.
\newblock {\em International Conference on Learning Representations}, 2016.

\bibitem[\protect\citeauthoryear{Lipton \bgroup \em et al.\egroup
  }{2016}]{lipton2016}
Zachary~C. Lipton, Xiujun Li, Jianfeng Gao, Lihong Li, Faisal Ahmed, and
  Li~Deng.
\newblock Efficient dialogue policy learning with bbq-networks.
\newblock {\em ArXiv: 1608.05081}, 2016.

\bibitem[\protect\citeauthoryear{Mnih \bgroup \em et al.\egroup
  }{2013}]{mnih2013}
Volodymyr Mnih, Koray Kavukcuoglu, David Silver, Alex Graves, Ioannis
  Antonoglou, Daan Wierstra, and Martin Riedmiller.
\newblock Playing atari with deep reinforcement learning.
\newblock {\em NIPS workshop in Deep Learning}, 2013.

\bibitem[\protect\citeauthoryear{Moerland \bgroup \em et al.\egroup
  }{2017}]{thomas2017}
Thomas~M. Moerland, Joost Broekens, and Catholijn~M. Jonker.
\newblock Efficient exploration with double uncertain value networks.
\newblock {\em Symposium on Deep Reinforcement Learning, NIPS}, 2017.

\bibitem[\protect\citeauthoryear{Morimura \bgroup \em et al.\egroup
  }{2010}]{morimura2010}
Tetsuro Morimura, Masashi Sugiyama, Hisashi Kashima, Hirotaka Hachiya, and
  Toshiyuki Tanaka.
\newblock Nonparametric return distribution approximation for reinforcement
  learning.
\newblock {\em ICML}, 2010.

\bibitem[\protect\citeauthoryear{Morimura \bgroup \em et al.\egroup
  }{2012}]{morimura2012}
Tetsuro Morimura, Masashi Sugiyama, Hisashi Kashima, Hirotaka Hachiya, and
  Toshiyuki Tanaka.
\newblock Parametric return density estimation for reinforcement learning.
\newblock {\em UAI}, 2012.

\bibitem[\protect\citeauthoryear{Osband and {Van Roy}}{2015}]{osband2015}
Ian Osband and Benjamin {Van Roy}.
\newblock Bootstrapped thompson sampling and deep exploration.
\newblock {\em arXiv:1507:00300}, 2015.

\bibitem[\protect\citeauthoryear{Osband \bgroup \em et al.\egroup
  }{2013}]{osband2013}
Ian Osband, Daniel Russo, and Benjamin {Van Roy}.
\newblock (more) efficient reinforcement learning via posterior sampling.
\newblock {\em Arxiv: 1306.0940}, 2013.

\bibitem[\protect\citeauthoryear{Osband \bgroup \em et al.\egroup
  }{2016}]{osband2016}
Ian Osband, Charles Blundell, Alexander Pritzel, and Benjamin {Van Roy}.
\newblock Deep exploration via bootstrapped dqn.
\newblock {\em arXiv:1602.04621}, 2016.

\bibitem[\protect\citeauthoryear{Osband \bgroup \em et al.\egroup
  }{2017}]{osband2017}
Ian Osband, daniel Russo, Zheng Wen, and Benjamin {Van Roy}.
\newblock Deep exploration via randomized value functions.
\newblock {\em arXiv: 1703.07608}, 2017.

\bibitem[\protect\citeauthoryear{Plappert \bgroup \em et al.\egroup
  }{2016}]{plappert2016}
Matthias Plappert, Rein Houthooft, Prafulla Dhariwal, Szymon Sidor, Richard~Y.
  Chen, Xi~Chen, Tamim Asfour, Pieter Abbeel, and Marcin Andrychowicz.
\newblock Parameter space noise for exploration.
\newblock {\em International Conference on Learning Representation}, 2016.

\bibitem[\protect\citeauthoryear{Ranganath \bgroup \em et al.\egroup
  }{2014}]{blei2015}
Rejesh Ranganath, Sean Gerrish, and David~M. Blei.
\newblock Black box variational inference.
\newblock {\em Proceedings of the 17th International Conference on Artificial
  Intelligence and Statistics (AISTATS)}, 2014.

\bibitem[\protect\citeauthoryear{Russo}{2017}]{russo2017}
Daniel Russo.
\newblock Tutorial on thompson sampling.
\newblock {\em arxiv}, 2017.

\bibitem[\protect\citeauthoryear{Schulman \bgroup \em et al.\egroup
  }{2015}]{schulman2015}
John Schulman, Sergey Levine, Philipp Moritz, Michael~I. Jordan, and Pieter
  Abbeel.
\newblock Trust region policy optimization.
\newblock {\em International Conference on Machine Learning}, 2015.

\bibitem[\protect\citeauthoryear{Tang and Kucukelbir}{2017}]{tang2017}
Yunhao Tang and Alp Kucukelbir.
\newblock Variational deep q network.
\newblock {\em 2nd Workshop on Bayesian Deep Learning, NIPS}, 2017.

\bibitem[\protect\citeauthoryear{Thompson}{1933}]{william1933}
William~R. Thompson.
\newblock On the likelihood that one unknown probability exceeds another in
  view of the evidence of two samples.
\newblock {\em Biometrika, Vol. 25, No. 3/4}, 1933.

\bibitem[\protect\citeauthoryear{Todorov \bgroup \em et al.\egroup
  }{2012}]{todorov2012}
Emanuel Todorov, Tom Erez, and Yuval Tassa.
\newblock Mujoco: A physics engine for model-based control.
\newblock {\em International Conference on Intelligent Robots}, 2012.

\bibitem[\protect\citeauthoryear{Tsitsiklis and {Van
  Roy}}{1996}]{tsitsiklis1996}
John~N. Tsitsiklis and Benjamin {Van Roy}.
\newblock Feature based methods for large scale dynamic programming.
\newblock {\em Machine Learning}, 1996.

\end{thebibliography}

\end{document}